\documentclass[sigconf]{acmart}
\usepackage{balance}
\usepackage{soul}
\usepackage{url}
\usepackage{wrapfig}
\usepackage{subfigure}
\def\model{GERBIL}
\AtBeginDocument{%
  }

\copyrightyear{2024}
\acmYear{2024}
\setcopyright{acmlicensed}\acmConference[CIKM '24]{Proceedings of the 33rd ACM International Conference on Information and Knowledge Management}{October 21--25, 2024}{Boise, ID, USA}
\acmBooktitle{Proceedings of the 33rd ACM International Conference on Information and Knowledge Management (CIKM '24), October 21--25, 2024, Boise, ID, USA}
\acmDOI{10.1145/3627673.3680041}
\acmISBN{979-8-4007-0436-9/24/10}

\begin{document}

\title{Revolutionizing Biomarker Discovery: \\ Leveraging Generative AI for Bio-Knowledge-Embedded Continuous Space Exploration}

\author{Wangyang Ying}
\orcid{0009-0009-6196-0287}
\email{wangyang.ying@asu.edu}
\affiliation{%
  \institution{Arizona State University}
  \city{Tempe}
  \state{Arizona}
  \country{USA}
}

\author{Dongjie Wang}
\orcid{0000-0003-3948-0059}
\email{wangdongjie@ku.edu}
\affiliation{%
  \institution{The University of Kansas}
  \city{Lawrence}
  \state{Kansas}
  \country{USA}}

\author{Xuanming Hu}
\orcid{0009-0002-2215-3553}
\email{solomonhxm@asu.edu}
\affiliation{%
  \institution{Arizona State University}
  \city{Tempe}
  \state{Arizona}
  \country{USA}
}

\author{Ji Qiu}
\email{Ji.Qiu@asu.edu}
\orcid{0000-0002-7913-9042}
\affiliation{%
  \institution{Arizona State University}
  \city{Tempe}
  \state{Arizona}
  \country{USA}
}

\author{Jin Park}
\orcid{0000-0001-6481-9590}
\email{Jin.Park.1@asu.edu}
\affiliation{%
 \institution{Arizona State University}
 \city{Tempe}
 \state{Arizona}
 \country{USA}}

\author{Yanjie Fu\textsuperscript{\textdagger}}
\orcid{0000-0002-1767-8024}
\email{yanjie.fu@asu.edu}
\affiliation{%
 \institution{Arizona State University}
 \city{Tempe}
 \state{Arizona}
 \country{USA}}
\thanks{\textsuperscript{\textdagger} Corresponding Author}

\renewcommand{\shortauthors}{Wangyang Ying et al.}

\begin{abstract}
Biomarker discovery is vital in advancing personalized medicine, offering insights into disease diagnosis, prognosis, and therapeutic efficacy. 
Traditionally, the identification and validation of biomarkers heavily depend on extensive experiments and statistical analyses. 
These approaches are time-consuming, demand extensive domain expertise, and are constrained by the complexity of biological systems.
These limitations motivate us to ask: \textit{Can we automatically identify the effective biomarker subset without substantial human efforts?}
Inspired by the success of generative AI, we think that the intricate knowledge of biomarker identification can be compressed into a continuous embedding space, thus enhancing the search for better biomarkers.
Thus, we propose a new biomarker identification framework with two important modules:1) training data preparation and 2) embedding-optimization-generation.
The first module uses a multi-agent system to automatically collect pairs of biomarker subsets and their corresponding prediction accuracy as training data.
These data establish a strong knowledge base for biomarker identification.
The second module employs an encoder-evaluator-decoder learning paradigm to compress the knowledge of the collected data into a continuous space. Then, it utilizes gradient-based search techniques and autoregressive-based reconstruction to efficiently identify the optimal subset of biomarkers.
Finally, we conduct extensive experiments on three real-world datasets to show the efficiency, robustness, and effectiveness of our method. The code is available at \url{http://tinyurl.com/bioDis}
\end{abstract}




\maketitle

\section{Introduction}
Within the biomedical field, Nucleic Acid Programmable Protein Array (NAPPA) technology is a critical resource that allows researchers and physicians to identify early illness biomarkers by simply drawing blood samples from patients. These biomarkers, such as protein components or antibodies, significantly improve the accuracy of treatment outcome predictions and contribute to the reduction of healthcare expenditures. 
NAPPA can identify a wide range of biomarkers owing to the extensive structural diversity of proteins. However, obtaining samples is often challenging. The resulting dataset falls into the category of classical high-dimensional and low-sample size (HDLSS) data~\cite{HDLSS1,HDLSS2,HDLSS3,HDLSS4}. However, the conventional biological methods for identifying a wide array of biomarkers are not only labor-intensive but also incur substantial costs for both healthcare providers and patients. Consequently, we propose employing machine learning to automatically identify a biomarker subset, aiming to reduce the dimensionality of this HDLSS data and achieve more effective predictive outcomes compared to traditional statistical methods.

Feature selection techniques are crucial in handling high dimensional data by identifying the optimal feature subset. To improve disease prediction, we recommend utilizing feature selection methods to discover crucial biomarkers. The feature selection methods fall into three categories:
1) Filter methods~\cite{kbest,forman2003extensive,hall1999feature,yu2003feature} select the top k features based on specific scores, often derived from univariate statistical tests. However, a drawback is that these methods are no-learnable, and statistical-based approaches might lack precision.
2) Embedded methods~\cite{lasso,sugumaran2007feature} simultaneously optimize feature selection and prediction tasks. However, embedded methods rely on strong structural assumptions and downstream models, imposing limitations on their flexibility.
3) Wrapper methods~\cite{yang1998feature,kim2000feature,narendra1977branch,kohavi1997wrappers} formulate feature selection as a search problem within a discrete feature combination space, often employing evolutionary algorithms with downstream models. However, these methods encounter challenges due to the exponential growth of the discrete search space (e.g., approximately $2^N$ for $N$ features). To address these issues, we propose a generative model perspective, aiming to avoid large, ineffective discrete searches and effectively identify the optimal biomarker subset.

\textbf{Our perspective: Biomarker Identification as Sequential Generative AI Task.} Emerging Artificial General Intelligence (AGI) and models like ChatGPT demonstrate the feasibility of learning complex and mechanistically unknown knowledge from historical experiences, and making wise decisions through autoregressive generation. Following a similar spirit, we believe that knowledge related to biomarkers can also be extracted and embedded into a continuous space, where computation and optimization are activated, and biomarker identification decisions are subsequently generated. This generative perspective treats biomarker identification (e.g., $b_1b_2,...,b_n \rightarrow b_1b_2b_4b_6$) as a sequential generation learning task to produce autoregressive biomarker identification decision sequences. Under this generative perspective, a biomarker subset is represented as a token sequence, which is then embedded into a differentiable continuous space. In this continuous space, each embedding vector corresponds to a biomarker subset, allowing us to: 1) construct an evaluation function to assess the utility of biomarker subsets; 2) search for the optimal biomarker subset embedding; and c) decode the embedding vector, generating biomarker token sequence.

Inspired by these findings, we propose a \textbf{deep variational sequential \underline{G}en\underline{ER}ative \underline{B}iomarker \underline{I}dentification \underline{L}earning (GERBIL)} framework, which includes two key components: training data preparation and embedding-optimization-generation.
Regarding the first component, we employ a multi-agent system to
collect training data as the biomarker identification knowledge base.
Specifically, for each biomarker, we create an agent to assess the appropriateness of its selection.
Then, the selected biomarker subset is used for predicting the disease status of each patient, with prediction accuracy serving as feedback to guide the next biomarker identification iteration.
The optimization objective of this process is to enhance the accuracy of disease status prediction.
Throughout this procedure, pairs of biomarker subset and prediction accuracy pairs are collected as the training data, encapsulating extensive knowledge on biomarker identification.
Regarding the second component,
it includes three steps:
\emph{1) Embedding.}
We have developed a variational transformer-based structure, jointly optimizing sequence reconstruction loss, biomarker subset utility evaluator loss, and variational distribution alignment (i.e., KL) loss to learn the embedding space of biomarker subsets. This strategy enhances the model's denoising capability, reducing the generation of noise features.
\emph{2) Optimization.}
Upon convergence of the embedding space, we leverage the evaluator to generate gradient and directional information, allowing us to guide gradient-based searches and identify the embedding of the optimal biomarker subset.
\emph{3) Generation.}
We decode the optimal embedding and autoregressively generate the sequence of optimal biomarkers. 
This optimal subset of biomarkers is anticipated to precisely predict patient status.
Finally, we conduct extensive experiments on three real-world datasets to validate the effectiveness of the proposed framework.
\section{Preliminaries and Problem Statement}

\noindent\textbf{Biomarker Token Sequence.}
To construct a differential embedding space for biomarkers, we need to collect $N$ biomarker subset-utility pairs as training data.
These data is denoted by $R = (\mathbf{b}_i, v_i)^N_{i=1}$, where $\mathbf{b}_i = [b_1, b_2, ..., b_q]$ is the biomarker token sequence of the $i$-th biomarker subset, and $v_i$ is corresponding downstream predictive utility.

\noindent\textbf{Problem Statement.}
Our task is to utilize biomarkers (a.k.a, antibodies) detected through NAPPA to predict whether a given sample is positive (belongs to a case group) for a certain disease. Our downstream task is to employ a random forest model to predict whether the sample presents a positive result.
Formally, consider a biological dataset $D = (B, y)$, where $B$ is an original biomarker set and y is the target label (case group or control group corresponding with patients). We collect the biomarker token sequences and their corresponding utilities by conducting automated biomarker identification on $D$.
Our goal is to 1) embed the knowledge of $R$ into a differentiable continuous space and 2) generate the optimal biomarker subset to classify the patients better. Regarding goal 1, we learn an encoder $\phi$, an evaluator $\vartheta$, and a decoder $\psi$ via joint optimization to get the embedding space $\mathcal{E}$. Regarding goal 2, we search for the best embedding and generate the optimal biomarker token sequences $\mathbf{b}^*$:
\begin{equation}
    \mathbf{b}^* = \psi(E^*) = \arg\max_{E \in \mathcal{E}}\mathcal{M}(B[\psi(E)], y), 
\end{equation}
where $\psi$ is a decoder to generate a biomarker token sequence from any embedding of $\mathcal{E}$; $E^*$ is the optimal biomarker subset embedding; $\mathcal{M}$ is a downstream ML task. We apply $\mathbf{b}^*$ to $B$ to select the optimal biomarker subset $B[\mathbf{b}^*]$.
\section{Methodology}

\subsection{Framework Overview}

\begin{figure*}[]
        \centering
	\includegraphics[width=1.0\linewidth]{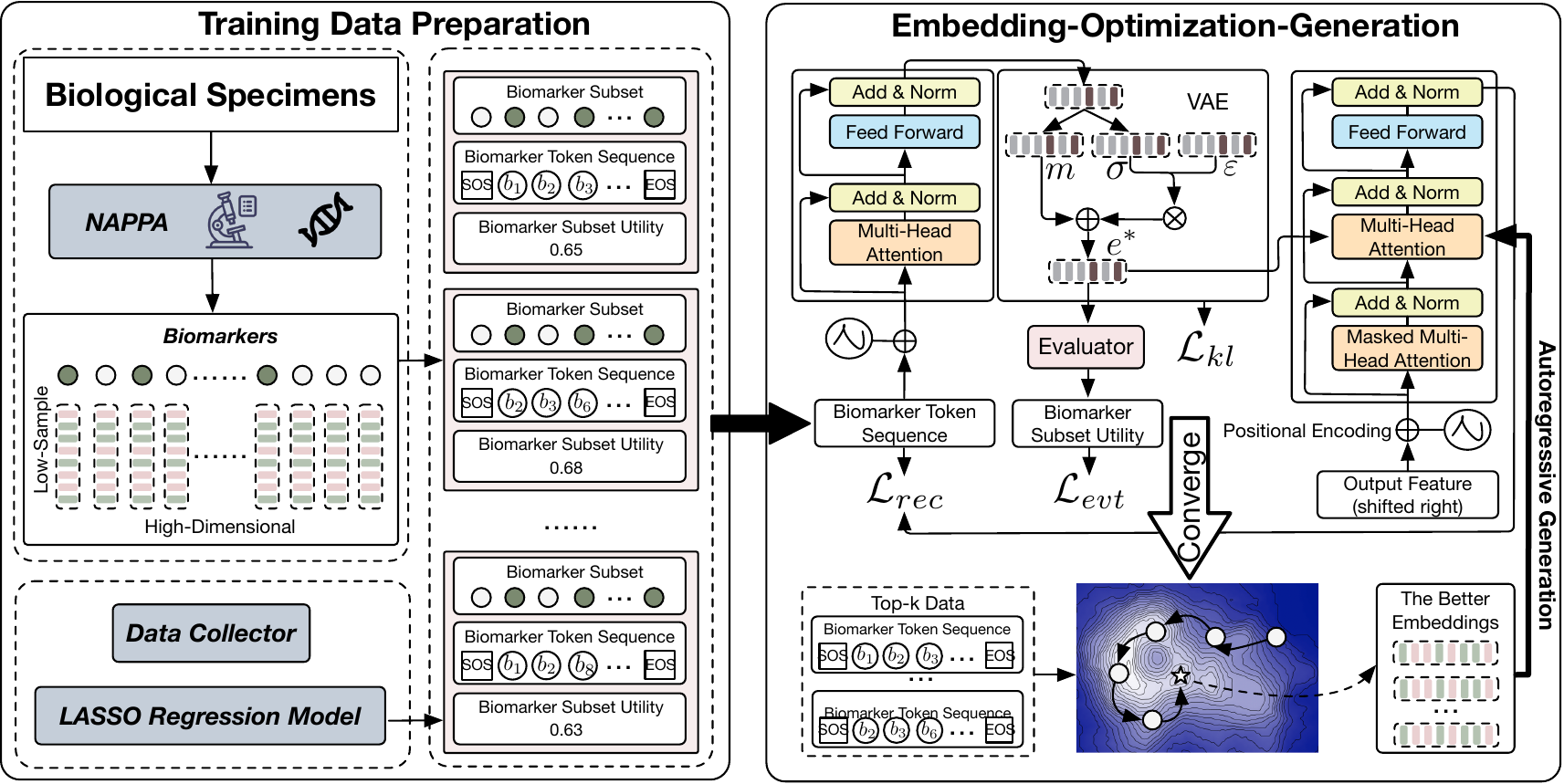}
	\caption{An overview of \model. First, we employ an RL-based data collector to collect biomarker subset-utility training data pairs. Second, we embed biomarker subsets into a continuous space through an embedding-optimization-generation paradigm. We search for the best embedding along the gradient direction maximizing the utility and generate the optimal biomarker subset based on the better embedding.}
        \vspace{-0.3cm}
	\label{framework}
\end{figure*}

Figure~\ref{framework} illustrates our method, which consists of two components:
1) training data preparation,
2) creation of the embedding-optimization-generation structure.
Specifically, step 1 aims to obtain historical biomarker identification experience (biomarker subsets) and their corresponding utility from high-dimensional and low-sample size biomarkers as training data. Due to the time-consuming nature of manually collecting training data, we leverage the automation and exploration of reinforcement learning to develop a biomarker subset data collector.
In step 2, we develop an embedding-optimization-generation paradigm to embed the knowledge of biomarker identification into a continuous space, and then identify the best biomarker subset. To achieve this, we develop an encoder-decoder-evaluator framework. Each biomarker is treated as a token, and a biomarker subset is considered a token sequence. The encoder encodes the biomarker token sequence into an embedding vector; the evaluator estimates the utility of the corresponding biomarker subset based on the embedding vector, and the decoder reconstructs the embedding vector into the respective biomarker token sequence. To build a distinguishable and smooth embedding space, we employed a variational transformer as the backbone of the sequential model, jointly optimizing sequence reconstruction loss and utility estimation loss to learn such an embedding space.
Then, we perform a gradient-guided search in the constructed embedding space to find better embedding vectors. We select the top k subsets based on the utility of the biomarker subset from the collected data, encode them into embedding vectors using the well-trained encoder, and then move these vectors in the direction of maximizing the biomarker subset utility using the gradient information provided by the well-trained evaluator.
Finally, we input the better vectors into the well-trained decoder to generate the biomarker token sequences. These sequences are applied to the original biomarker set to generate biomarker subsets. We use random forest to test the generated biomarker subsets, and the subset with the highest performance is the optimal result.

\subsection{Training Data Preparation}
\noindent\textbf{High-dimensional and Low-sample Size Biomarkers.}
NAPPA can effectively identify antibodies in biological specimens used to distinguish whether the specimen contains a particular disease; these antibodies are referred to as biomarkers. However, due to the diversity of proteins, the number of detected biomarkers is often extensive, and biological specimens are typically challenging to obtain, for example, in rare diseases where positive specimens are scarce, or due to ethical constraints making specimen acquisition difficult. The collected biomarkers exhibit the typical characteristics of high-dimensional and low-sample size. Such data not only has a small sample size but also may have highly collinear biomarkers (i.e., linear correlation). Biomarkers unrelated to the label may lead to identification errors and risks of model overfitting. Therefore, dimensionality reduction of biomarkers is essential to identify key biomarkers for distinguishing the respective diseases.

\noindent\textbf{Biomarker Identification Knowledge Acquisition.}
To embed the knowledge of biomarker identification into a continuous space and then facilitate the identification of the best biomarker subsets within this space, we require two essential components as training data: 1) historical biomarker identification experience, and 2) the corresponding utility values associated with these biomarker subsets.

Inspired by~\cite{marlfs}, we propose that the identification of biomarker subsets can be effectively modeled through a multi-agent system. 
To implement this concept, we introduce an automated data collector system based on reinforcement learning, specifically utilizing a reinforced agent (DQN~\cite{DQN}) for each biomarker. Each agent has two actions: selecting or deselecting the corresponding biomarker, with the representation of the chosen biomarker subset serving as the state of the agent. We employ a random forest model as the downstream machine learning model to evaluate the utility of the identified biomarker subsets, and the utility is used to give feedback to agents as a reward. This system operates iteratively, collaborating among multiple agents to select the biomarker subset.
During each iteration, the chosen biomarker subset is input into a downstream machine learning model to obtain the associated utility. The overarching optimization objective is to maximize the performance of the downstream machine learning model while minimizing redundancy in the selected biomarker subset.
The iterative exploration process of this system facilitates the collection of a substantial volume of data samples.

\noindent\textbf{Biomarker token sequences with shuffling-based augmentations.} 
To effectively construct the biomarker subset embedding space, we treat each biomarker subset as a biomarker token sequence. These sequences can be encoded into embedding vectors by a sequential model. 
We observe that the utility of the biomarker token sequence remains unaffected by its order. Exploiting this observation, we introduce a shuffling-based strategy aimed at rapidly expanding our pool of valid data samples
For instance, give one sample ``$b_1, b_2, b_3$"$\rightarrow0.867$, we can shuffle the order of the sequence to generate more semantically equivalent data samples: ``$b_2,b_1,b_3$"$\rightarrow0.867$, ``$b_3,b_2,b_1$"$\rightarrow0.867$. 
The shuffling augmentation strategy enhances both the volume and diversity of data, enabling the construction of an empirical training set that more accurately represents the true population. This strategy is significant in developing a more effective continuous embedding space.

\subsection{Embedding-Optimization-Generation}
The success of ChatGPT showcases the effectiveness of embedding complex human knowledge in a vast space through sequential modeling. This success motivates the incorporation of biomarker identification, a form of human knowledge, into a continuous embedding space. 
Our goal is not just to preserve biomarker subset knowledge in this space but also to maintain their utility, crucial for identifying optimal subsets. To achieve this, we propose a novel learning paradigm with an encoder-decoder-evaluator framework.

\noindent\textbf{Embedding: Construction of the biomarker subset embedding space via variational transformer.}
We develop an encoder-decoder-evaluator paradigm for embedding biomarker identification knowledge into a continuous space.
This space is designed to retain the impact of various biomarker subsets, while also possessing a smooth structure to facilitate the identification of optimal embeddings. 
To accomplish this, we adopt the variational transformer~\cite{vaswani2017attention,vae} as the backbone for our sequential model, providing a robust foundation for the implementation of this structure.

\underline{\textit{The Encoder}} aims to embed a biomarker token sequence into an embedding vector.  
Formally, consider a training dataset $R = {(\mathbf{b}_i, v_i)}_{i=1}^N $, where $\mathbf{b}_i$ and $v_i$ are a biomarker token sequence and corresponding utility of the $i$-th training data respectively, and $N$ is the number of samples. To simplify the notation, we use $(\mathbf{b}, v)$ to represent any training data. 
We first employ a transformer encoder $\phi$  to learn the embedding of the biomarker token sequence, denoted by  $\mathbf{e} = \phi(\mathbf{b})$. 
We assume that the learned embeddings $\mathbf{e}$ follow the format of normal distribution.
Then, two fully connected layers are implemented to estimate the mean $\mathbf{m}$ and variance $\mathbf{\sigma}$ of this distribution.
After that, we can sample an embedding vector $\mathbf{e^*}$ from the distribution via the reparameterization technique.
This process is denoted by $\mathbf{e^*} = \mathbf{m} + \mathbf{\varepsilon} * exp(\mathbf{\sigma})$, where $\varepsilon$ refers to the noised vector sampled from a standard normal distribution.
The sampled vector $\mathbf{e^*}$ is regarded as the input of the following decoder and evaluator.

\underline{\textit{The Decoder}} aims to reconstruct a biomarker token sequence using the embedding $\mathbf{e^*}$.  
We utilize a transformer decoder to parse the information of $\mathbf{e^*}$ and add a softmax layer behind it to estimate the probability of the next biomarker token based on the previous ones.
Formally, consider $\mathbf{b} = [b_1, b_2,...,b_q]$, where $q$ represents the length of the biomarker token sequence. The current token that needs to be decoded is $b_p$, and the previously completed biomarker token sequence is $b_1...b_{p-1}$. 
The probability of the $p$-th token should be:
$
    P_\psi(b_p | \mathbf{e^*}, [b_1,b_2,...,b_{p-1}]) = \frac{exp(z_p)}{\sum_q exp(z)},
$
where $z_p$ represents the $p$-th output of the softmax layer, $\psi$ refers to the decoder.
The joint estimated likelihood of the entire biomarker token sequence should be: 
$
    P_\psi(\mathbf{b} | \mathbf{e^*}) = \prod_{p=1}^q P_\psi(b_p | \mathbf{e^*},[b_1,b_2,...,b_{p-1}])
$

\underline{\textit{The Evaluator}} aims to evaluate the biomarker subset utility based on the embedding $\mathbf{e^*}$.
More specifically, we implement a fully connected neural layer as the evaluator to predict the corresponding utility in the sequential training data.
This calculation process can be denoted by $\ddot{v} = \vartheta(\mathbf{e^*})$, where $\vartheta$ refers to the evaluator and $\ddot{v}$ is the predicted utility via $\vartheta$.

\underline{\textit{The Joint Optimization.}} 
We jointly train the encoder, decoder, and evaluator to learn the continuous embedding space. 
There are three objectives: 
a) Minimizing the reconstruction loss between the reconstructed biomarker token sequence and the real one, denoted by
$
    \mathcal{L}_{rec} = -\sum_{p=1}^qlogP_\psi(b_p | \mathbf{e^*}, [b_1,b_2,...,b_{p-1}]),
$
b) Minimizing the estimation loss between the predicted utility and the real one, denoted by
$
    \mathcal{L}_{evt} = MSE(v, \ddot{v}),
$
c) Minimizing the Kullback–Leibler (KL) divergence between the learned distribution of the biomarker subset and the standard normal distribution, denoted by
$
    \mathcal{L}_{kl} = \sum(exp(\sigma_i) - (1 + \sigma_i) + (m_i)^2).
$
The first two objectives ensure that each point within the embedding space is associated with a specific biomarker subset and its corresponding predictive utility. The last objective smoothens the embedding space, thereby enhancing the efficacy of the following gradient-steered search step.

\noindent\textbf{Optimization: Gradient-guided search for the best biomarker subset embedding.}
After obtaining the biomarker subset embedding space,  we employ a gradient-ascent search 
method to find better biomarker subset embedding. 
More specifically, we first select the top K biomarker subset from the collected data based on the corresponding utility.
Then, we utilize the well-trained encoder to convert these subsets as the local optimal embeddings.
After that, we adopt a gradient-ascent algorithm to move these embeddings along the direction maximizing the downstream predictive accuracy.
The gradient comes from the well-trained evaluator $\vartheta$.
Taking the embedding $\mathbf{e^*}$ as an example, the moving calculation process is as follows:
$
    \mathbf{e^+} = \mathbf{e^*} +\eta\frac{\partial\vartheta}{\partial{\mathbf{e^*}}},
$
where $\eta$ is the moving steps and $\mathbf{e^+}$ is the better embedding. 
\setlength{\tabcolsep}{2.3mm}{
\begin{table*}[tb]
\centering
\fontsize{7.5}{7}\selectfont
\caption{Overall Performance. This table highlights the best results by \textbf{bold} fonts. We evaluate the performance of \model\ and the baselines regarding precision, recall, F-1 score, and AUC. The higher the value is, the better the biomarker subset quality is.}
\begin{tabular}{c|cccc|cccc|cccc}
\toprule
\toprule
Dataset  & \multicolumn{4}{c|}{GC}                                           & \multicolumn{4}{c|}{EBVaGC}                                       & \multicolumn{4}{c}{IM}                                      \\ \midrule
         & Precision       & Recall         & F-1 Score       & AUC            & Precision       & Recall         & F-1 Score       & AUC            & Precision & Recall         & F-1 Score       & AUC            \\ \midrule
Original & 0.427          & 0.430          & 0.425          & 0.430          & 0.595          & 0.564          & 0.534          & 0.557          & 0.551    & 0.550          & 0.544          & 0.549          \\
F-test   & 0.746          & 0.740          & 0.737          & 0.740          & 0.770          & 0.756          & 0.744          & 0.750          & 0.750    & 0.740          & 0.736          & 0.740          \\
mRMR     & 0.808          & 0.800          & 0.798          & 0.790          & 0.743          & 0.742          & 0.734          & 0.733          & 0.761    & 0.750          & 0.747          & 0.750          \\
MCDM     & 0.458          & 0.469          & 0.458          & 0.470          & 0.412          & 0.420          & 0.408          & 0.410          & 0.530    & 0.530          & 0.523          & 0.530          \\
RFE      & 0.790          & 0.760          & 0.752          & 0.760          & 0.780          & 0.758          & 0.750          & 0.748          & 0.784    & \textbf{0.780} & 0.777          & \textbf{0.780} \\
LASSO    & 0.644          & 0.639          & 0.639          & 0.639          & 0.535          & 0.565          & 0.540          & 0.550          & 0.559    & 0.550          & 0.532          & 0.550          \\
LASSONet & 0.516          & 0.520          & 0.503          & 0.520          & 0.672          & 0.646          & 0.637          & 0.641          & 0.573    & 0.570          & 0.563          & 0.570          \\
GFS      & 0.667          & 0.659          & 0.654          & 0.660          & 0.649          & 0.647          & 0.642          & 0.639          & 0.713    & 0.710          & 0.709          & 0.710          \\
MARLFS   & 0.499          & 0.500          & 0.489          & 0.500          & 0.677          & 0.647          & 0.623          & 0.634          & 0.651    & 0.640          & 0.631          & 0.640          \\
SARLFS   & 0.497          & 0.490          & 0.483          & 0.490          & 0.644          & 0.625          & 0.565          & 0.599          & 0.619    & 0.610          & 0.599          & 0.610          \\ \midrule
\model   & \textbf{0.850} & \textbf{0.840} & \textbf{0.839} & \textbf{0.840} & \textbf{0.879} & \textbf{0.854} & \textbf{0.846} & \textbf{0.840} & \textbf{0.785}    & \textbf{0.780} & \textbf{0.779} & \textbf{0.780} \\ \bottomrule \bottomrule
\end{tabular}
\vspace{-0.3cm}
\label{exp:overall_performance}
\end{table*}}

\noindent\textbf{Generation: Autoregressive generation of the best biomarker subset.}
Once we identify the better embeddings, we proceed to decode the biomarker token sequences based on them in an autoregressive manner.
Formally, we take the embedding $\mathbf{e^+}$ as an example to illustrate the decoding process. In the $p$-iteration, we assume that the previously generated biomarker token sequence is $b_1...b_{p-1}$ and the waiting to generate token is $b_p$.
The estimation probability for generating $b_q$ is to maximize the following likelihood based on the well-trained decoder $\psi$: $b_p = \arg\max(P_\psi(b_p | \mathbf{e^+}, [b_1,...,b_{p-1}])$.
We will iteratively generate the possible biomarker tokens until finding the end token (i.e., $<$EOS$>$).
For instance, if the generated token sequence is ``$[b_2,b_6,b_5,\text{$<$EOS$>$}, b_8]$, '', we will cut from the \text{$<$EOS$>$} token and keep 
$[b_2, b_5, b_6]$ as the final generation result.
Finally, we select the corresponding biomarkers according to these tokens and output the biomarker subset with the highest utility as the optimal biomarker subset.
\section{Experiments}
\subsection{Experimental Settings}
\noindent\textbf{Data Description.}
We conduct experiments on three real-world biological datasets:
1) \textbf{Gastric Cancer (GC)}~\cite{GC}: GC dataset evaluated humoral responses to a nearly complete H. pylori immunoproteome using NAPPA. This dataset includes 3,440 biomarkers, 50 GC cases, and 50 GC controls.
2) \textbf{Epstein–Barr virus-associated Gastric Cancer (EBVaGC)}~\cite{EBVaGC}: EBVaGC dataset characterized the GC-specific antibody response to EBV, which detects EBV-positive GC and elucidates its contribution to carcinogenesis. This dataset includes 3,440 biomarkers, 28 EBV-positive cases, and 34 EBV-negative controls. 
3) \textbf{Intestinal Metaplasia (IM)}~\cite{IM}: IM dataset evaluated humoral responses to H. pylori proteins among IM and non-atrophic gastritis using H. pylori protein arrays. This dataset includes 3,448 biomarkers, 50 IM gastritis cases, and 50 non-atrophic gastritis controls.

\noindent\textbf{Evaluation Metrics.}
We use a random forest (RF) model to evaluate the performance of the identified biomarker subset because RF is stable and robust, and can reduce the prediction variation caused by downstream models. We used the 5-fold cross-validation to evaluate the performance of our method and baseline algorithms in terms of precision, recall, F-1 score, and AUC.

\noindent\textbf{Reproducibility.}
1) Data Collector: We use the reinforcement data collector to explore 500 epochs to collect feature subset-utility data pairs, and randomly shuffle each feature sequence 25 times to augment the training data.
2) Feature Subset Embedding: We map feature tokens to a 64-dimensional embedding, and use a 2-layer network for both encoder and decoder, with a multi-head setting of 8 and a feed-forward layer dimension of 256. The latent dimension of the VAE is set to 64. The estimator consists of a 2-layer feed-forward network, with each layer having a dimension of 200. The values of $\alpha$, $\beta$, and $\gamma$ are 0.8, 0.2, and 0.001, respectively. We set the batch size as 1024, the training epochs as 400, and the learning rate as 0.0001. 
3) Optimal Embedding Search and Reconstruction: We use the top 25 feature sets to search for the feature subsets and keep the optimal feature subset. 

\noindent\textbf{Baseline Algorithms.}
We compared our method with 9 widely used baselines:
(A) Filter methods: 1) \textbf{F-test}~\cite{F-tests} select the top-$k$ biomarkers with the highest important scores;
2) \textbf{mRMR}~\cite{mrmr} selects a biomarker subset by maximizing relevance with labels and minimizing feature-feature redundancy; 
3) \textbf{MCDM}~\cite{mcdm} ensemble biomarker identification as a Multi-Criteria Decision-Making problem, which uses the VIKOR sort algorithm to rank features based on the judgment of multiple feature selection methods;
(B) Embedding methods:
4) \textbf{RFE}~\cite{rfe} recursively deletes the weakest biomarkers; 
5) \textbf{LASSO}~\cite{lasso} shrinks the coefficients of useless biomarkers to zero by sparsity regularization to select features; 
6) \textbf{LASSONet}~\cite{lassonet} is a neural network with sparsity to encourage the network to use only a subset of input biomarkers; 
(C) Wrapper methods:
7) \textbf{GFS}~\cite{fan2021autogfs} is a group-based biomarker identification method via interactive reinforcement learning; 
8) \textbf{MARLFS}~\cite{marlfs} uses reinforcement learning to create an agent for each biomarker to learn a policy to select or deselect the corresponding biomarker, and treat biomarker redundancy and downstream task performance as rewards; 
9) \textbf{SARLFS}~\cite{sarlfs} is a simplified version of MARLFS to leverage a single agent to replace multiple agents to decide the actions of all biomarkers.
To evaluate the necessity of each technical component of \model, we develop two model variants:
i) \textbf{\model$^+$} removes the variational component and solely uses the Transformer to create the feature subset embedding space;
ii) \textbf{\model$^{-}$} adopts LSTM~\cite{LSTM} to learn the feature subset embedding space.
\subsection{Experimental Results}

\noindent\textbf{Overall Performance.}
In this experiment, we evaluate the performance of \model\ and baseline algorithms for biomarker identification across three real-world biological datasets in terms of precision, recall, F1 score, and AUC. Table~\ref{exp:overall_performance} illustrates that \model\ consistently outperforms other baselines across all datasets, showcasing significant performance improvements over baseline models and original datasets. The underlying driver of this observation is the ability of \model\ to compress biomarker identification knowledge into an extensive embedding space. Such compression facilitates a more effective search for optimal biomarker identification results.

\begin{figure}[]
	\centering
	\subfigure[GC]{\label{exp:vae:GC}\includegraphics[width=0.155\textwidth]{{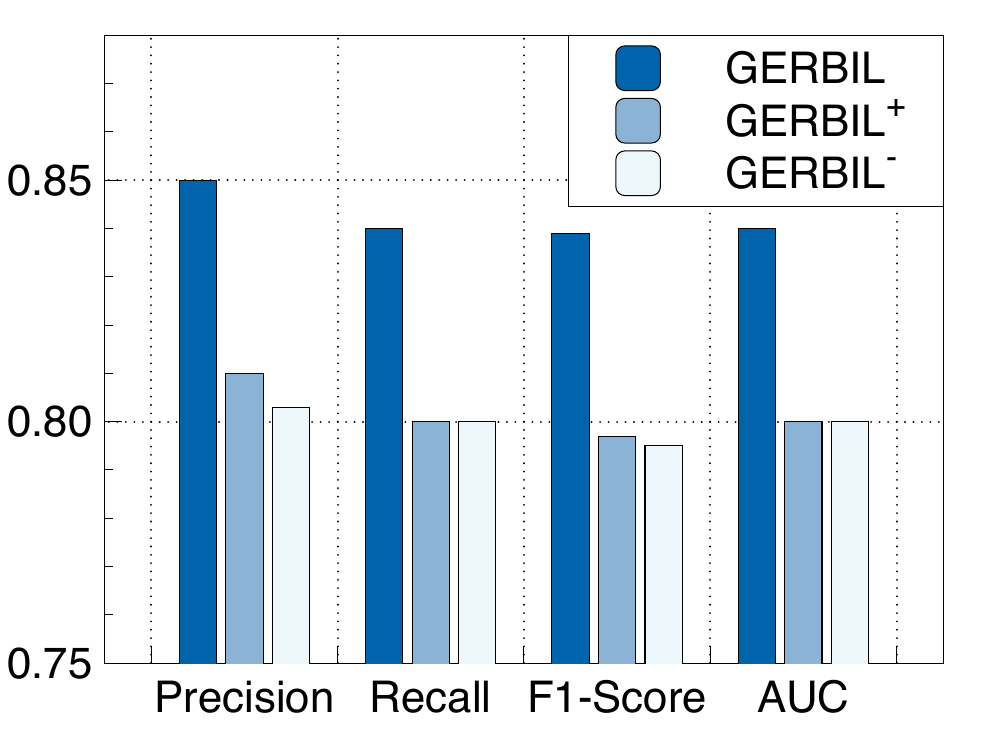}}}
	\subfigure[EBVaGC]{\label{exp:vae:EBVaGC}\includegraphics[width=0.155\textwidth]{{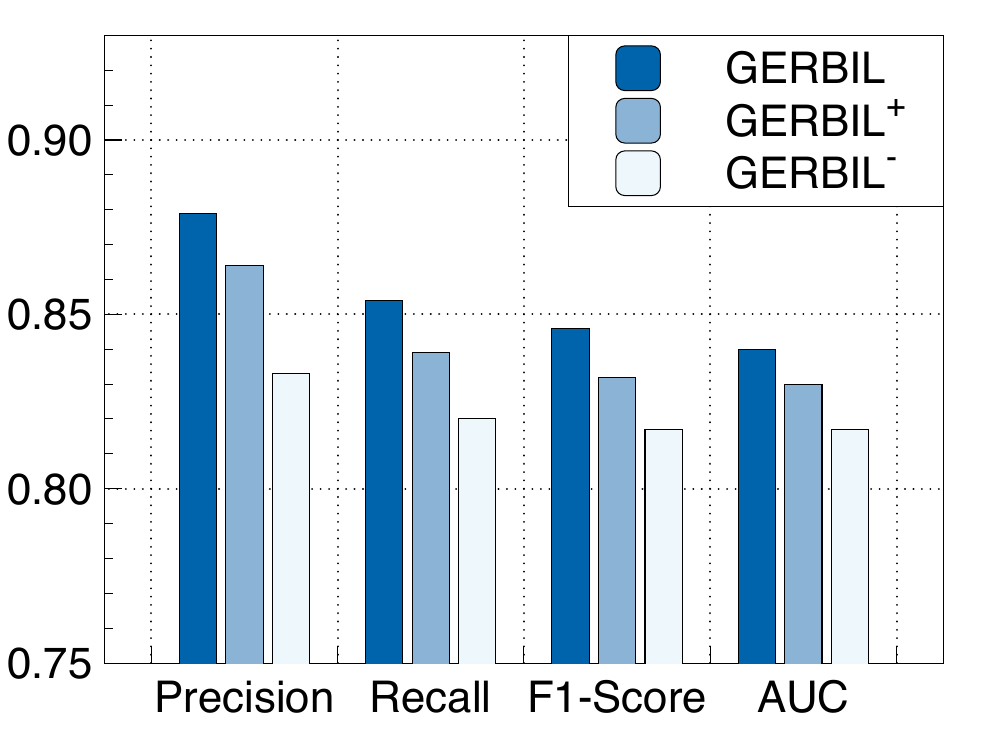}}}
         \subfigure[IM]{\label{exp:vae:IM}\includegraphics[width=0.155\textwidth]{{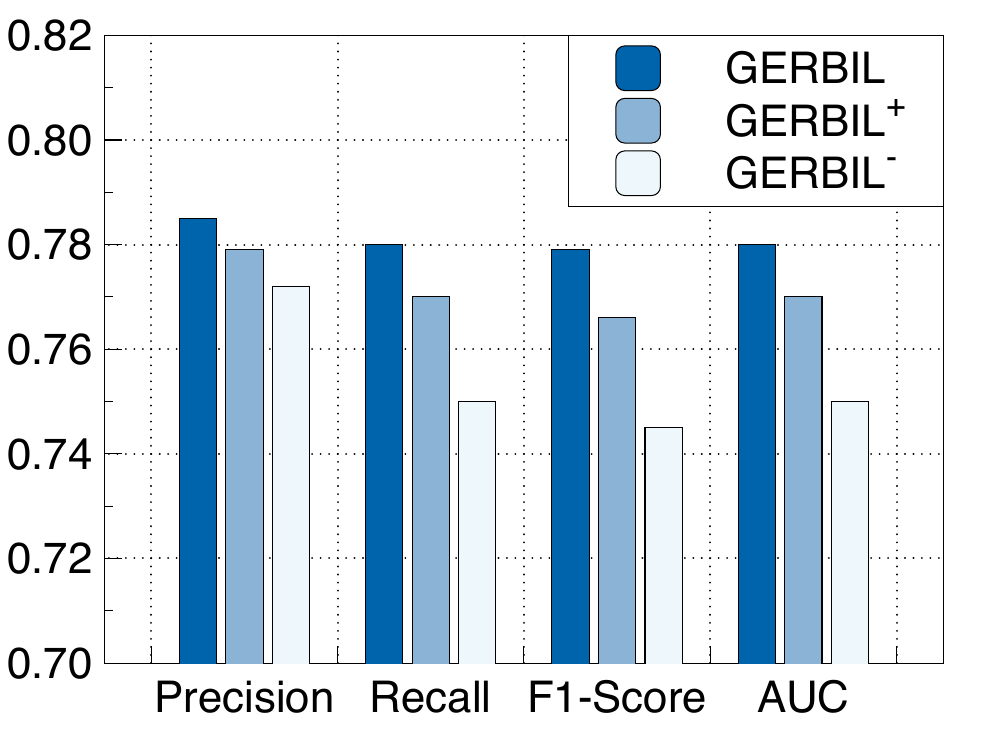}}}
         \vspace{-0.3cm}
	\caption{Analysis of the impact of variational transformer.}
	\label{exp:ablation_vae}
 \vspace{-0.3cm}
\end{figure}

\noindent\textbf{The impact of the variational transformer for biomarker identification.}
One key aspect of \model\ is its utilization of a sequential model to embed biomarker identification knowledge into an embedding space. To analyze the impact of the sequential model choice, we developed two model variants: 1) \model$^+$; 2) \model$^-$. Figure~\ref{exp:ablation_vae} demonstrates that \model\ outperforms \model$^+$ across all datasets. The potential reason lies in the enhancement of the smoothness of the embedded space for learned biomarker subsets in \model\ due to the variational component. This smoothness contributes to a more effective search for optimal biomarker identification results. Additionally, another intriguing observation is that \model$^+$ outperforms \model$^-$ across all datasets. One potential explanation for this observation is that, compared to LSTM, the transformer architecture excels in capturing complex correlations between different biomarker combinations and their impact on the performance of downstream machine learning tasks. In summary, this experiment highlights the necessity of each technical component in \model.
\begin{figure}[]
	\centering
	\subfigure[GC]{\label{exp:rl:GC}\includegraphics[width=0.155\textwidth]{{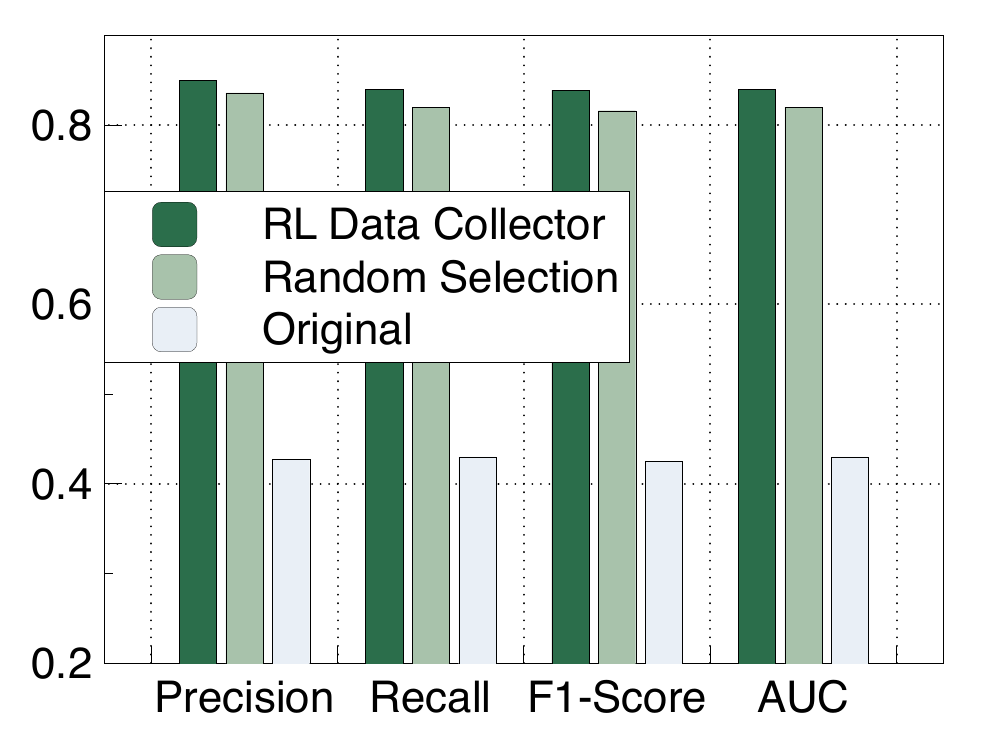}}}
	\subfigure[EBVaGC]{\label{exp:rl:EBVaGC}\includegraphics[width=0.155\textwidth]{{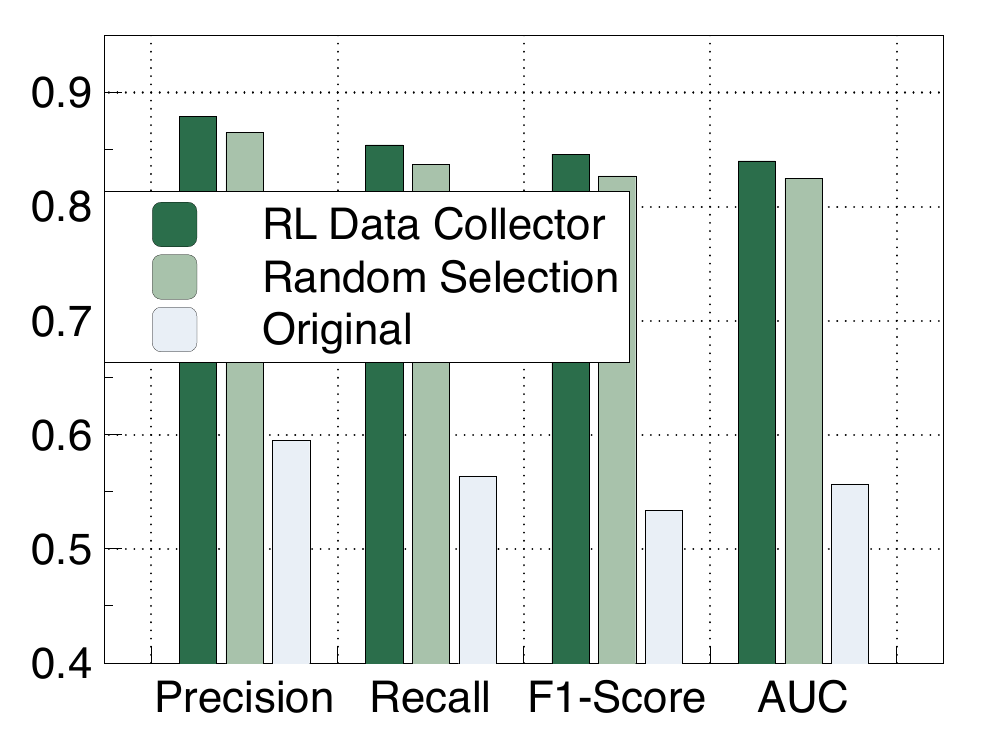}}}
         \subfigure[IM]{\label{exp:rl:IM}\includegraphics[width=0.155\textwidth]{{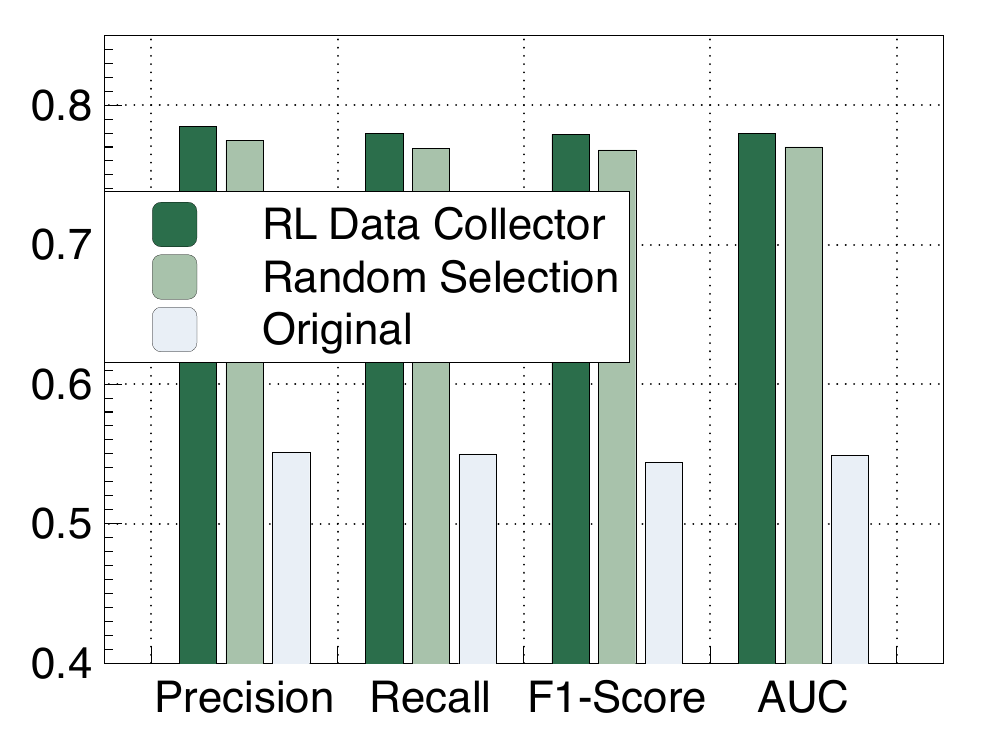}}}
         \vspace{-0.3cm}
	\caption{Analysis of the impact of RL data collector.}
	\label{exp:ablation_rl}
 \vspace{-0.3cm}
\end{figure}

\noindent\textbf{The impact of the RL-based data collector.}
In \model, we emphasize the capability of the RL-based data collector to gather higher-quality training data, thereby facilitating the construction of a better embedding space. To assess the impact of the RL-based data collector, we established two control groups: a) randomly collecting data samples to construct the embedding space; b) directly using the original dataset for prediction. Figure~\ref{exp:ablation_rl} shows that the data collected using the RL-based collector can identify biomarker subsets superior to both control groups. The underlying driver is that the RL-based collector can procure higher-quality data, contributing to the creation of a more effective embedding space. This enhanced embedding space facilitates the search for better biomarker subsets. Another observation is that even when constructing the embedding space using randomly collected data and subsequently searching for the optimal subset, the performance in downstream tasks significantly improves compared to the original biomarker set. This suggests that \model, even based on randomly collected data, can learn biomarker knowledge, thereby substantially enhancing the final identification results.

\begin{figure}[]
	\centering
	\subfigure[GC]{\label{exp:aug:GC}\includegraphics[width=0.155\textwidth]{{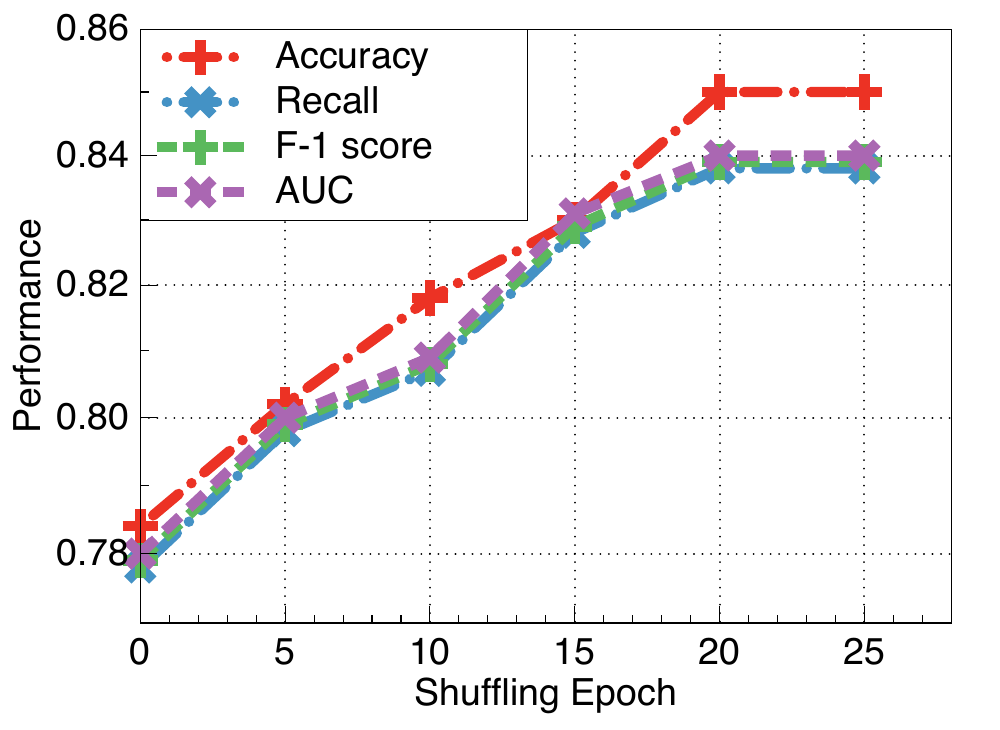}}}
	\subfigure[EBVaGC]{\label{exp:aug:EBVaGC}\includegraphics[width=0.155\textwidth]{{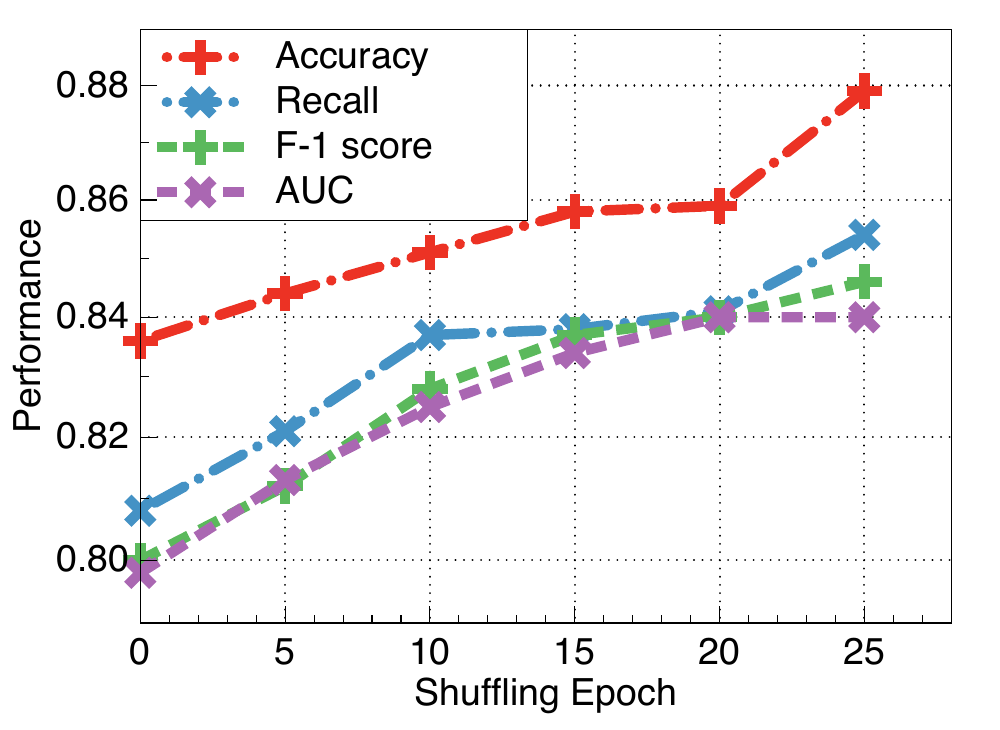}}}
         \subfigure[IM]{\label{exp:aug:IM}\includegraphics[width=0.155\textwidth]{{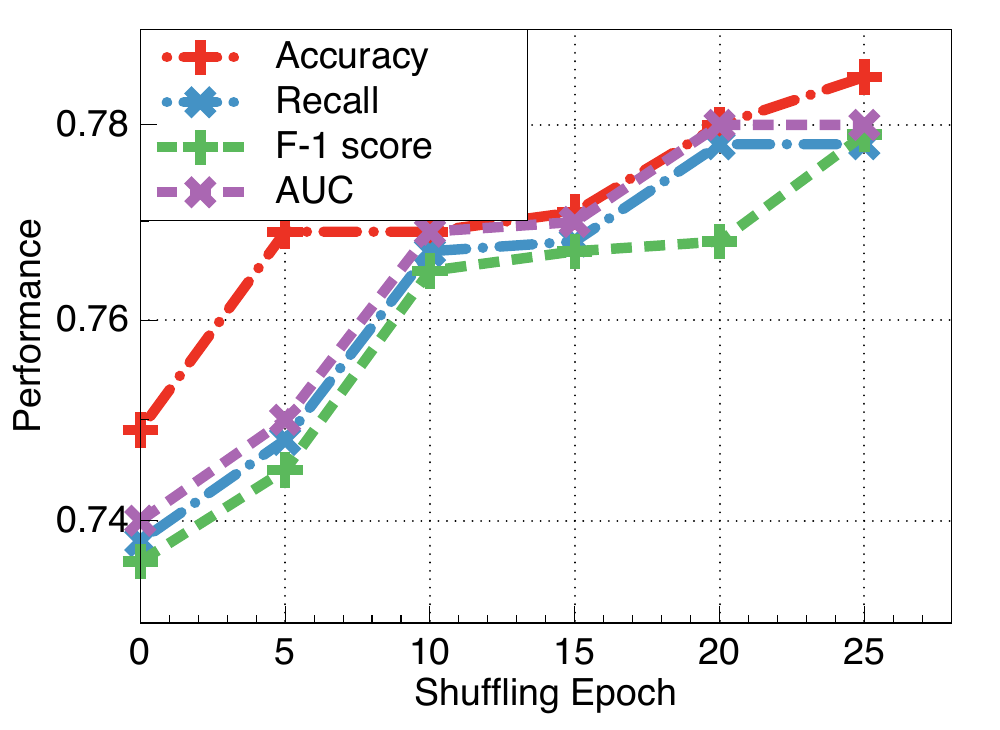}}}
         \vspace{-0.3cm}
	\caption{Analysis of the impact of data augmentation.}
	\label{exp:ablation_aug}
 \vspace{-0.3cm}
\end{figure}

\noindent\textbf{The impact of the shuffling augmentation.}
Since the order of the biomarker token sequence does not impact the performance of the sequence, \model\ employs a random shuffle of sequences as data augmentation. In this experiment, we explore the influence of data augmentation on our final identification results. From Figure~\ref{exp:ablation_aug}, we observe that with an increase in the number of shuffling iterations, downstream machine learning performance improves across all datasets. One potential reason is that a higher number of shuffling iterations enhances the diversity and volume of the data, thereby strengthening the construction of the embedding space. The improved embedding space possesses better generation, leading to superior biomarker identification results
\begin{figure}[]
	\centering
	\subfigure[Training]{\label{exp:complexity:train}\includegraphics[width=0.155\textwidth]{{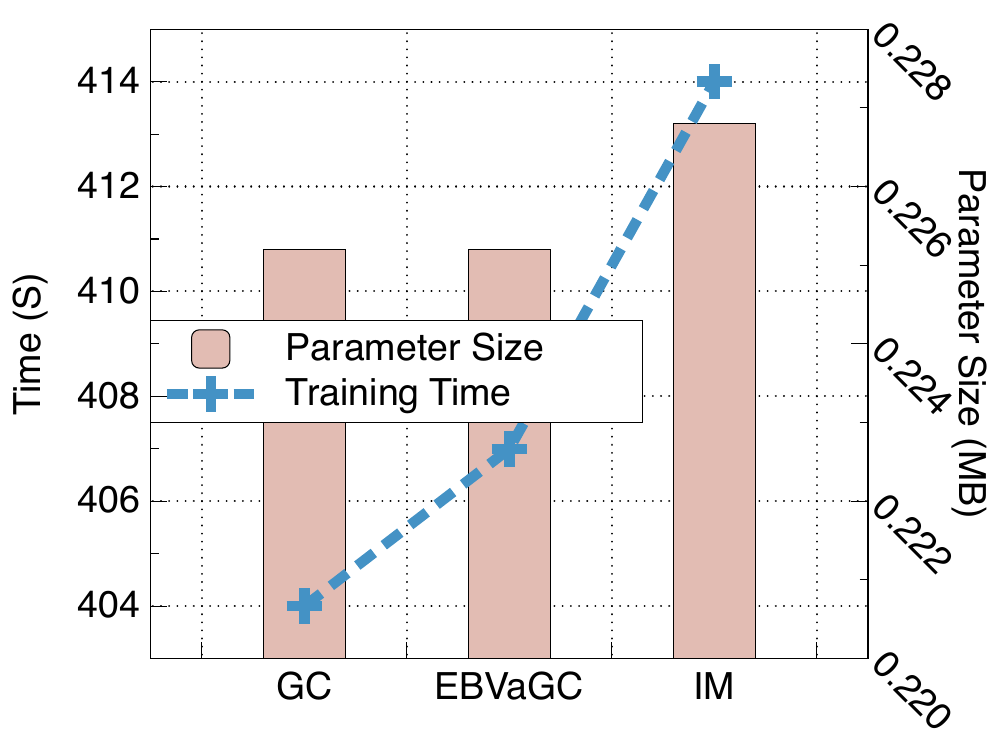}}}
	\subfigure[Inference]{\label{exp:complexity:infer}\includegraphics[width=0.155\textwidth]{{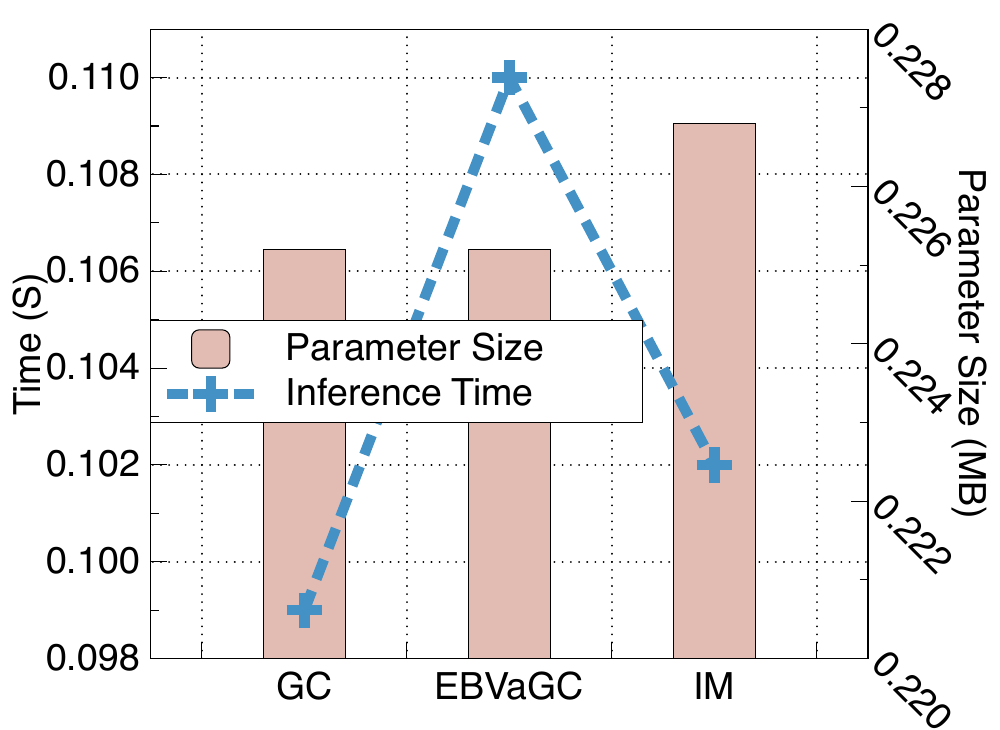}}}
         \subfigure[Data Collection]{\label{exp:complexity:collect}\includegraphics[width=0.155\textwidth]{{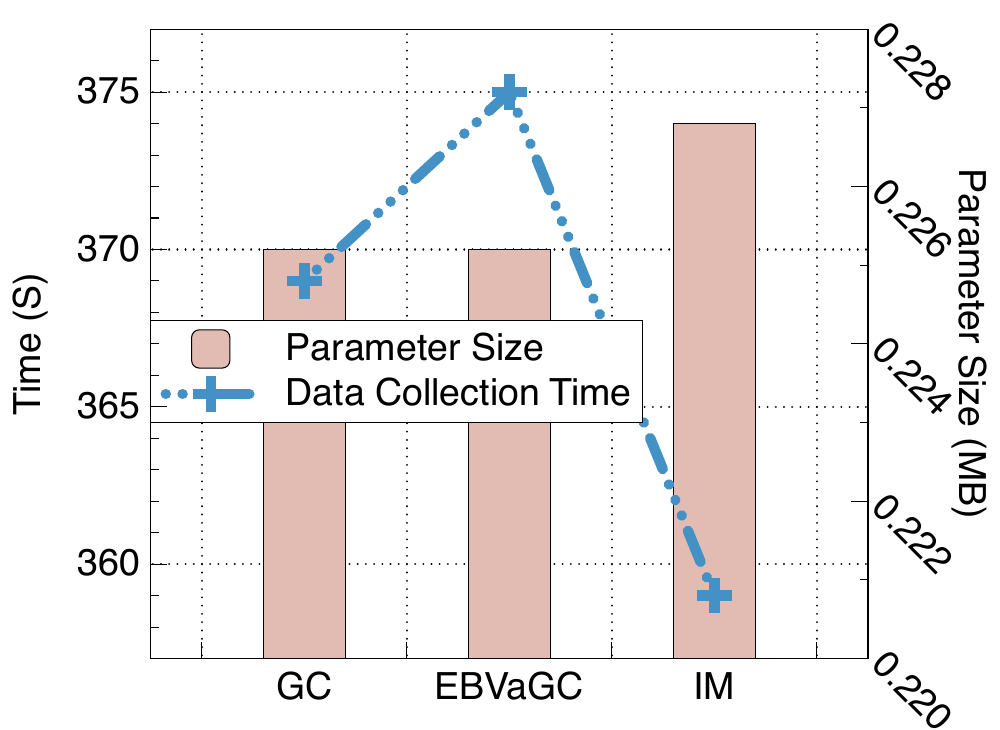}}}
         \vspace{-0.3cm}
	\caption{Time and space complexity check.}
	\label{exp:ablation_complexity}
 \vspace{-0.3cm}
\end{figure}

\noindent\textbf{Time and space complexity of \model.}
This experiment reports on the evaluation of the time and space complexity of the \model, considering data collection time, model training time, model inference time, and model size. Figure~\ref{exp:ablation_complexity} indicates that the model has a small number of parameters and can complete data collection and model training in a short period across all datasets. One possible explanation is that \model can rapidly capture knowledge of biomarkers, leading to quick convergence. Another observation is that once the model converges, inference time significantly decreases, benefiting from mapping the sequences to a low-dimensional space, enabling fast inference. This experiment demonstrates the efficiency of the \model\ on HDLLS data.

\noindent\textbf{Robustness check.}
To assess the robustness of different biomarker identification algorithms under various downstream machine learning (ML) models, we replaced the random forest model with decision trees (DT), XGBoost (XGB), support vector machines (SVM), and logistic regression (LR) to evaluate algorithm performance on three datasets. The comparative results are presented in Table~\ref{exp:robustness_check}. We observe that \model\ consistently achieves the best or second-best results, irrespective of the downstream ML model employed. The underlying driver is that \model\ can customize biomarker identification strategies based on the specific biomarkers of the downstream ML model. This is achieved by collecting sequentially trained data tailored to each model type. Additionally, \model\ embeds biomarker identification knowledge into a continuous embedding space, enhancing its robustness and generalization across different ML models. In conclusion, this experiment indicates that \model\ can maintain its outstanding and stable biomarker identification performance across different ML models.
\setlength{\tabcolsep}{1mm}{
\begin{table*}[tb]
\centering
\fontsize{7}{7}\selectfont
\caption{Robustness Check. In this table, the best results and the second results are highlighted by \textbf{bold} and \underline{underlined} fonts respectively. We evaluate the performance of \model\ and all baselines on different downstream tasks. The higher the value is, the better the performance is.}
\begin{tabular}{@{}cc|cccc@{}}
\toprule \toprule
                                              & Model                & DT                                     & XGB                                    & SVM                                    & LR                                     \\ \cmidrule(l){3-6} 
Dataset                                       &                      & Precision , Recall , F-1 score , AUC    & Precision , Recall , F-1 score , AUC    & Precision , Recall , F-1 score , AUC    & Precision , Recall , F-1 score , AUC    \\ \midrule
\multicolumn{1}{c|}{}                         & F-test               & 0.520 , 0.520 , 0.514 , 0.520          & 0.757 , 0.750 , 0.746 , 0.750          & 0.629 , 0.600 , 0.588 , 0.600          & \textbf{0.805 , 0.780 , 0.774 , 0.780} \\
\multicolumn{1}{c|}{}                         & mRMR                 & \underline{0.685 , 0.680 , 0.678 , 0.680}    & 0.709 , 0.700 , 0.697 , 0.700          & \textbf{0.843 , 0.830 , 0.828 , 0.830} & 0.726 , 0.660 , 0.638 , 0.660          \\
\multicolumn{1}{c|}{}                         & MCDM                 & 0.480 , 0.480 , 0.478 , 0.480          & 0.437 , 0.440 , 0.435 , 0.440          & 0.518 , 0.520 , 0.515 , 0.520          & 0.476 , 0.470 , 0.451 , 0.470          \\
\multicolumn{1}{c|}{}                         & RFE                  & 0.675 , 0.650 , 0.644 , 0.650          & \textbf{0.777} , \underline{0.760 , 0.755 , 0.760}    & 0.644 , 0.630 , 0.623 , 0.630          & 0.640 , 0.630 , 0.613 , 0.630          \\
\multicolumn{1}{c|}{{GC}}                     & LASSO                & 0.523 , 0.520 , 0.515 , 0.520          & 0.551 , 0.550 , 0.548 , 0.550          & 0.798 , 0.790 , 0.784 , 0.790    & 0.736 , 0.710 , 0.699 , 0.710          \\
\multicolumn{1}{c|}{}                         & LASSONet             & 0.439 , 0.440 , 0.428 , 0.440          & 0.499 , 0.500 , 0.493 , 0.500          & 0.456 , 0.460 , 0.429 , 0.460          & 0.525 , 0.520 , 0.510 , 0.520          \\
\multicolumn{1}{c|}{}                         & GFS                  & 0.644 , 0.640 , 0.636 , 0.640          & 0.599 , 0.600 , 0.589 , 0.600          & 0.585 , 0.570 , 0.542 , 0.570          & 0.567 , 0.610 , 0.575 , 0.610          \\
\multicolumn{1}{c|}{}                         & MARLFS               & 0.561 , 0.560 , 0.558 , 0.560          & 0.550 , 0.550 , 0.545 , 0.550          & 0.635 , 0.630 , 0.626 , 0.630          & 0.642 , 0.630 , 0.623 , 0.630          \\
\multicolumn{1}{c|}{}                         & SARLFS               & 0.465 , 0.470 , 0.464 , 0.470          & 0.461 , 0.460 , 0.457 , 0.460          & 0.605 , 0.600 , 0.595 , 0.600          & 0.655 , 0.650 , 0.646 , 0.650          \\
\multicolumn{1}{c|}{}                         & \textbf{\model}      & \textbf{0.725 , 0.710 , 0.706 , 0.710} & \underline{0.776} , \textbf{0.760 , 0.756 , 0.760}        & \underline{0.838 , 0.830 , 0.828 , 0.830}          & \underline{0.743 , 0.720 , 0.710 , 0.720}    \\ \midrule
\multicolumn{1}{c|}{}                         & F-test               & 0.674 , 0.659 , 0.656 , 0.660          & \underline{0.808 , 0.787 , 0.785 , 0.791}    & \underline{0.741 , 0.738 , 0.721 , 0.727}    & 0.668 , 0.644 , 0.618 , 0.622          \\
\multicolumn{1}{c|}{}                         & mRMR                 & 0.576 , 0.563 , 0.557 , 0.565          & 0.752 , 0.742 , 0.739 , 0.739          & 0.741 , 0.724 , 0.716 , 0.713          & \underline{0.698 , 0.678 , 0.662 , 0.658}    \\
\multicolumn{1}{c|}{}                         & MCDM                 & 0.457 , 0.447 , 0.447 , 0.448          & 0.548 , 0.549 , 0.531 , 0.540          & 0.444 , 0.485 , 0.427 , 0.469          & 0.302 , 0.549 , 0.389 , 0.500          \\
\multicolumn{1}{c|}{}                         & RFE                  & 0.676 , 0.663 , 0.658 , 0.665          & 0.741 , 0.724 , 0.716 , 0.713          & 0.716 , 0.694 , 0.684 , 0.684          & 0.494 , 0.596 , 0.476 , 0.553          \\
\multicolumn{1}{c|}{{EBVaGC}}                 & LASSO                & 0.504 , 0.503 , 0.493 , 0.506          & 0.474 , 0.469 , 0.455 , 0.468          & 0.586 , 0.596 , 0.577 , 0.580          & 0.623 , 0.597 , 0.573 , 0.575          \\
\multicolumn{1}{c|}{}                         & LASSONet             & 0.502 , 0.500 , 0.489 , 0.492          & 0.627 , 0.583 , 0.573 , 0.590          & 0.644 , 0.614 , 0.609 , 0.618          & 0.598 , 0.597 , 0.572 , 0.579          \\
\multicolumn{1}{c|}{}                         & GFS                  & \underline{0.711 , 0.708 , 0.695 , 0.702}    & 0.646 , 0.645 , 0.636 , 0.630          & 0.553 , 0.568 , 0.528 , 0.547          & 0.302 , 0.549 , 0.389 , 0.500          \\
\multicolumn{1}{c|}{}                         & MARLFS               & 0.534 , 0.536 , 0.531 , 0.532          & 0.435 , 0.471 , 0.444 , 0.460          & 0.556 , 0.535 , 0.510 , 0.529          & 0.493 , 0.504 , 0.488 , 0.497          \\
\multicolumn{1}{c|}{}                         & SARLFS               & 0.597 , 0.596 , 0.591 , 0.585          & 0.601 , 0.599 , 0.591 , 0.585          & 0.553 , 0.551 , 0.524 , 0.547          & 0.647 , 0.615 , 0.589 , 0.604          \\
\multicolumn{1}{c|}{}                         & \textbf{\model} & \textbf{0.780 , 0.777 , 0.776 , 0.772} & \textbf{0.815 , 0.792 , 0.788 , 0.793} & \textbf{0.759 , 0.725 , 0.715 , 0.712} & \textbf{0.745 , 0.726 , 0.715 , 0.716} \\ \midrule
\multicolumn{1}{c|}{}                         & F-test               & \underline{0.732 , 0.720 , 0.715 , 0.720}    & 0.747 , 0.740 , 0.737 , 0.740          & \textbf{0.786 , 0.750 , 0.741 , 0.750} & \textbf{0.714 , 0.700 , 0.691 , 0.700} \\
\multicolumn{1}{c|}{}                         & mRMR                 & 0.684 , 0.680 , 0.678 , 0.680          & \underline{0.750 , 0.740 , 0.736 , 0.740}    & 0.665 , 0.650 , 0.643 , 0.650          & 0.684 , 0.680 , 0.678 , 0.680          \\
\multicolumn{1}{c|}{}                         & MCDM                 & 0.445 , 0.450 , 0.446 , 0.450          & 0.541 , 0.540 , 0.537 , 0.540          & 0.566 , 0.560 , 0.557 , 0.560          & 0.488 , 0.460 , 0.449 , 0.460          \\
\multicolumn{1}{c|}{}                         & RFE                  & 0.734 , 0.730 , 0.730 , 0.730          & 0.742 , 0.720 , 0.715 , 0.720          & 0.665 , 0.650 , 0.643 , 0.650          & 0.704 , 0.680 , 0.670 , 0.680          \\
\multicolumn{1}{c|}{{IM}}                     & LASSO                & 0.562 , 0.560 , 0.555 , 0.560          & 0.554 , 0.550 , 0.546 , 0.550          & 0.670 , 0.660 , 0.655 , 0.660          & 0.679 , 0.670 , 0.665 , 0.670          \\
\multicolumn{1}{c|}{}                         & LASSONet             & 0.531 , 0.530 , 0.523 , 0.530          & 0.545 , 0.540 , 0.531 , 0.540          & 0.573 , 0.560 , 0.543 , 0.560          & 0.522 , 0.510 , 0.495 , 0.510          \\
\multicolumn{1}{c|}{}                         & GFS                  & 0.702 , 0.700 , 0.699 , 0.700          & 0.722 , 0.720 , 0.719 , 0.720          & 0.590 , 0.590 , 0.589 , 0.590          & 0.572 , 0.570 , 0.568 , 0.570          \\
\multicolumn{1}{c|}{}                         & MARLFS               & 0.539 , 0.540 , 0.539 , 0.540          & 0.606 , 0.600 , 0.591 , 0.600          & 0.610 , 0.610 , 0.600 , 0.610          & 0.642 , 0.640 , 0.637 , 0.640          \\
\multicolumn{1}{c|}{}                         & SARLFS               & 0.601 , 0.590 , 0.577 , 0.590          & 0.583 , 0.580 , 0.575 , 0.580          & 0.562 , 0.560 , 0.558 , 0.560          & 0.539 , 0.540 , 0.537 , 0.540          \\
\multicolumn{1}{c|}{}                         & \textbf{\model}                & \textbf{0.735, 0.730 , 0.728 , 0.730}  & \textbf{0.751 , 0.750 , 0.749 , 0.750} & \underline{0.683 , 0.670 , 0.663 , 0.670}    & \underline{0.684 , 0.682 , 0.679 , 0.680}    \\ \bottomrule \bottomrule
\end{tabular}
\vspace{-0.3cm}
\label{exp:robustness_check}
\end{table*}}
\section{Related Work}
In the field of biological data, filter methods~\cite{kbest,mrmr,mcdm}, especially univariate statistical tests, are widely applied in biomarker identification. These methods are computationally efficient for biomarker identification in high-dimensional data. The F-test~\cite{F-tests} is a common statistical method for biomarker identification, assessing the correlation between biomarkers and labels based on the statistical properties of the data and selecting the subset of biomarkers with the highest scores. Other classical statistical methods such as student's t-test~\cite{T-test}, Pearson correlation test~\cite{pearson}, Chi-square test~\cite{chi-square}, etc., can be similarly applied for biomarker identification. These methods have low computational complexity, allowing for the quick and effective identification of biomarker subsets from high-dimensional datasets. However, they overlook the dependencies and interactions between biomarkers, potentially leading to suboptimal results.
Wrapper methods~\cite{gfe,marlfs,sarlfs,ying2024feature,gong2024neuro,ning2024fedgcs}, based on specific datasets, predefine machine learning models, and iteratively evaluate candidate biomarker subsets. These methods often outperform filter methods as they assess the entire biomarker set. However, they require enumerating all possible biomarker subsets for evaluation, posing an NP-hard problem, especially in high-dimensional datasets where the computational cost is high, making it challenging to identify the optimal biomarker subset.
Embedded methods~\cite{lasso,lassonet,rfe,C2IMUFS,xiao2023beyond,zhang2024enhanced} transform the biomarker identification task into a regularization term in machine learning model prediction loss to accelerate the identification process. For example, the LASSO family methods, while capable of handling high-dimensional and low-sample size data, rely on L1 regularization and specific downstream tasks, limiting their applicability. Furthermore, the consideration of only linear relationships between biomarkers leads to suboptimal performance.
In comparison to the existing approaches mentioned above, we propose a novel generative AI perspective. This perspective embeds biomarker identification knowledge into a continuous embedding space and then employs gradient-guided search and autoregressive generation to effectively identify the optimal biomarker subset.
\section{Conclusion}
In biomarker discovery, we introduce a generative model to automatically identify the effective biomarker subset without human efforts.
There are three important contributions:
1) we propose a new formulation, which treats biomarker identification as a deep generative AI to covert the discrete biomarker identification process into a continuous optimization; 
2) we develop a multi-agent system to automatically collect biomarker subset knowledge, facilitating the construction of the biomarker subset embedding space;
3) we develop an embedding-optimization-generation paradigm to embed biomarker subset knowledge, facilitating the gradient-steered optimal embedding identification and the best biomarker subset generation.  
This structure enables the generation of optimal results, avoiding the need to explore exponentially growing possibilities of biomarker combinations in discrete space. Experiments on three real-world datasets highlight its potential as a valuable approach for biomarker discovery in the bioinformatics domain.

\begin{acks}
This research was partially supported by the National Science Foundation (NSF) via the grant numbers: 2421864, 2421803, 2421865, and National academy of Engineering Grainger Foundation Frontiers of Engineering Grants.
\end{acks}

\bibliographystyle{ACM-Reference-Format}
\balance
\bibliography{sample-base}

\end{document}